\documentclass[letterpaper]{article} 
\usepackage{aaai25}  
\usepackage{times}  
\usepackage{helvet}  
\usepackage{courier}  
\usepackage[hyphens]{url}  
\usepackage{graphicx} 
\usepackage{natbib}  
\usepackage{caption} 
\usepackage{algorithm}
\usepackage{algorithmic}
\usepackage{booktabs}
\usepackage{amsmath}
\usepackage{amssymb}
\usepackage{newfloat}
\usepackage{listings}
\DeclareCaptionStyle{ruled}{labelfont=normalfont,labelsep=colon,strut=off} 
\floatstyle{ruled}
\newfloat{listing}{tb}{lst}{}
\floatname{listing}{Listing}
\title{LLaVA Needs More Knowledge: Retrieval Augmented Natural Language Generation with Knowledge Graph for Explaining Thoracic Pathologies}
\author{
    Ameer Hamza\textsuperscript{\rm 1},
    Abdullah\textsuperscript{\rm 1},
    Yong Hyun Ahn\textsuperscript{\rm 2},
    Sungyoung Lee\textsuperscript{\rm 1},
    Seong Tae Kim\textsuperscript{\rm 1}\thanks{Corresponding author.}
}
\affiliations{
    \textsuperscript{\rm 1}Department of Computer Science and Engineering, Kyung Hee University, Republic of Korea\\
    \textsuperscript{\rm 2}Department of Artificial Intelligence, Kyung Hee University, Republic of Korea \\

    \{ameer, abdullahijaz, yhahn, sylee0301, st.kim\}@khu.ac.kr
}

\begin{document}

\maketitle

\begin{abstract}
Generating Natural Language Explanations (NLEs) for model predictions on medical images, particularly those depicting thoracic pathologies, remains a critical and challenging task. Existing methodologies often struggle due to general models' insufficient domain-specific medical knowledge and privacy concerns associated with retrieval-based augmentation techniques. To address these issues, we propose a novel Vision-Language framework augmented with a Knowledge Graph (KG)-based datastore, which enhances the model's understanding by incorporating additional domain-specific medical knowledge essential for generating accurate and informative NLEs. Our framework employs a KG-based retrieval mechanism that not only improves the precision of the generated explanations but also preserves data privacy by avoiding direct data retrieval. The KG datastore is designed as a plug-and-play module, allowing for seamless integration with various model architectures. We introduce and evaluate three distinct frameworks within this paradigm: KG-LLaVA, which integrates the pre-trained LLaVA model with KG-RAG; Med-XPT, a custom framework combining MedCLIP, a transformer-based projector, and GPT-2; and Bio-LLaVA, which adapts LLaVA by incorporating the Bio-ViT-L vision model. These frameworks are validated on the MIMIC-NLE dataset, where they achieve state-of-the-art results, underscoring the effectiveness of KG augmentation in generating high-quality NLEs for thoracic pathologies.
\end{abstract}

\begin{links}
\link{Code}{https://github.com/ailab-kyunghee/KG-LLaVA}
\end{links}

\section{Introduction}

In recent years, natural language processing (NLP) has witnessed the development of numerous models trained on vast amounts of general domain data  \cite{radford2019language, brown2020language, touvron2023llama, jiang2023mistral, peng2023instruction}
While these models exhibit remarkable capabilities across various tasks, they often lack the specialized knowledge required for domain-specific applications, such as generating Natural Language Explanations (NLEs) for thoracic pathologies. This limitation is particularly pronounced in the medical domain, where accurate and contextually relevant explanations are crucial for diagnostic decision-making.

To bridge this gap, including Pre-Training, Fine-Tuning, and Retrieval-Augmented Generation (RAG) methods has been explored. One popular approach among these strategies is Pre-Training large models on medical data.
However, this strategy necessitates extensive data and substantial computational resources \cite{saab2024capabilities, li2024llava}. In the medical domain, publicly available datasets are often limited, and concerns about data privacy, authenticity, and potential data leakage persist. Even when models are pre-trained on medical data, they frequently struggle with task-specific performance and may suffer from reduced factual accuracy. Moreover, pre-training these models is both expensive and time-consuming, and subsequent fine-tuning is often required to adapt them to specific downstream tasks.

Fine-tuning is another common strategy, where general domain models are adapted directly to medical tasks by training on specialized datasets. While this approach can be effective, it is hampered by the scarcity of high-quality medical datasets and the need to protect patient privacy and data security. Additionally, fine-tuned models can be prone to hallucination, generating explanations that lack factual correctness. Recent advancements in parameter-efficient fine-tuning, such as training low-rank adapters \cite{hu2021lora}, have aimed to reduce computational costs while maintaining model performance, yet challenges remain, particularly in maintaining the model's generalization across diverse tasks.

The third strategy, RAG \cite{lewis2020retrieval}, has gained traction as a method for enhancing general domain models with domain-specific knowledge dynamically. In this approach, models are fine-tuned on task-specific data while being supplemented with relevant information retrieved from a datastore. RAG methods have shown promising results in maintaining factual accuracy and reducing hallucination risks. However, the effectiveness of RAG is highly dependent on the quality of the retrieval mechanism. Also, in the medical domain, concerns about data privacy are amplified, as retrieved information might still be traceable to individual patients, even after de-identification, thus posing a risk of data leakage.

To overcome these challenges, we propose a novel approach that combines the strengths of vision-language models with a Knowledge Graph (KG)-based retrieval system. Our method, KG-based Retrieval-Augmented Generation (KG-RAG), addresses privacy risks by abstracting patient-specific details and providing models with more relevant and factual information tailored to individual cases. The KG-based datastore serves as a robust source of domain-specific knowledge, enabling the generation of accurate and contextually appropriate NLEs for thoracic pathologies.

KG-RAG emulates the cognitive process of radiologists, who rely on extensive experience and domain-specific knowledge to formulate diagnostic explanations. By leveraging a KG-based datastore tailored to each patient case, our approach significantly enhances the model's performance in generating precise and informative explanations. To demonstrate the versatility and effectiveness of our method, we integrated KG-RAG into three distinct frameworks: KG-LLaVA, Med-XPT, and Bio-LLaVA.

KG-LLaVA integrates the pre-trained LLaVA model with our KG-RAG module, fine-tuning it on our dataset to enrich its ability to generate detailed and accurate explanations by leveraging the CLIP ViT-L vision model. Med-XPT is a custom-built framework combining MedCLIP as the vision encoder, a transformer-based projector, and GPT-2 as the language model, trained from scratch on the MIMIC-NLE dataset to fully exploit the domain-specific knowledge provided by the KG-RAG module. Lastly, Bio-LLaVA adapts the LLaVA model by replacing the vision encoder with Bio-ViT-L, a model tailored for biomedical tasks, and modifying the projection layer to accommodate the unique feature dimensions of Bio-ViT-L. This framework was trained exclusively on the MIMIC-NLE dataset, showcasing its ability to generate precise NLEs without relying on pre-trained projector weights.

This integration of advanced vision-language models with a domain-specific KG not only provides transparent and comprehensible reasoning for detected abnormalities but also elevates the model's diagnostic accuracy. Our approach underscores the potential of combining state-of-the-art machine learning techniques with domain-specific knowledge to achieve expert-level reasoning, thereby improving the interpretability and accuracy of diagnostic outcomes in chest X-ray images.

Our main contributions can be summarized as follows:
\begin{itemize}
\item We propose the first KG retrieval-augmented Vision-Language Model (VLM) framework specifically designed for generating NLEs for thoracic pathologies. This approach integrates domain-specific medical knowledge into the explanation generation process, enhancing the accuracy and relevance of the outputs.

\item Our method addresses critical privacy concerns associated with medical data by abstracting patient-specific details through the use of a KG-based datastore. Furthermore, the proposed method is designed as a plug-and-play module, making it easily adaptable to existing radiology tasks and compatible with previous methods.

\item We validate the effectiveness of our approach by achieving state-of-the-art results on a benchmark dataset, MIMIC-NLE. Our method outperforms previous models, demonstrating the robustness and applicability of the KG-augmented framework in the medical domain.
 \end{itemize}

 \begin{figure*}[t]
\centering
\includegraphics[width=0.99\textwidth]{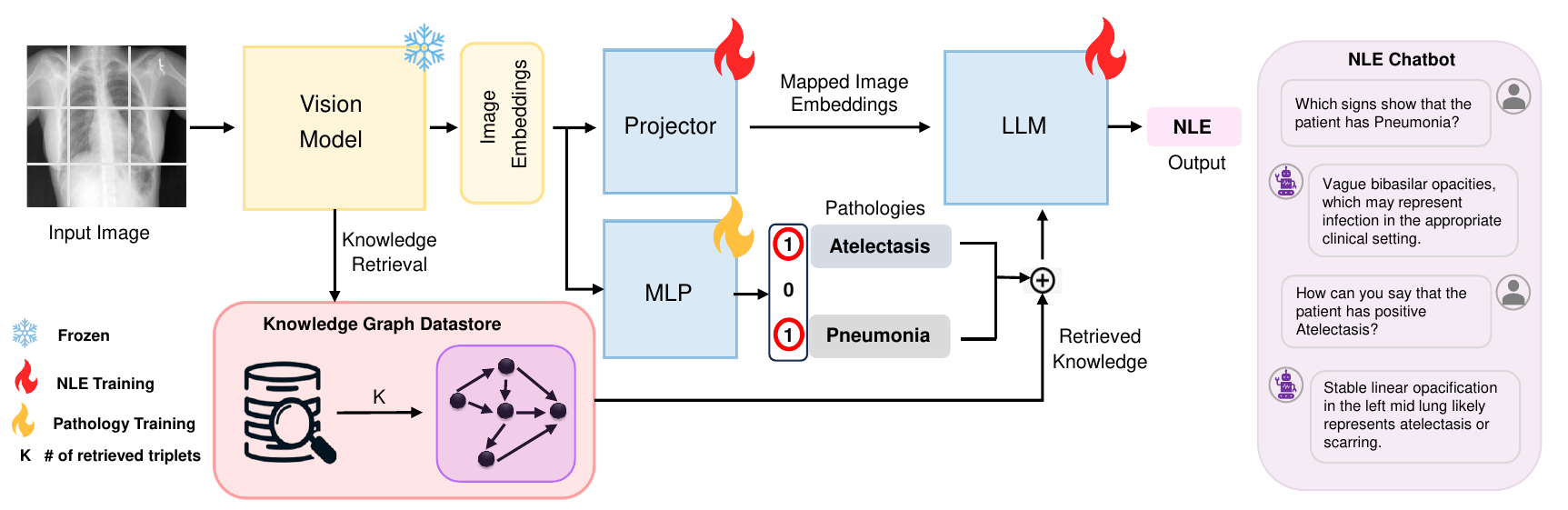} 
\caption{Overview of the KG-LLaVA framework with integrated Knowledge Graph Retrieval Augmented Generation (KG-RAG) module. The framework combines a pre-trained LLaVA model with a CLIP ViT-L vision encoder to extract visual features, which are then projected into the language model's embedding space. The KGR module uses MedCLIP to map input images to a shared latent space and retrieve relevant KG triplets via the FAISS library. These triplets provide domain-specific context that enhances the generation of accurate and informative NLEs for thoracic pathologies. The modular design allows for seamless integration with other architectures, such as Med-XPT and Bio-LLaVA, ensuring flexibility and adaptability across different vision-language tasks.}
\label{fig1}
\end{figure*}

\section{Related Work}
\subsubsection{Natural Language Explanation.} NLEs provide textual interpretations of deep learning model predictions, aiming to offer accessible and comprehensible insights for users, particularly in complex domains like medical diagnostics. \citeauthor{hendricks2016generating} was the first to introduce the NLE task. This task was later extended to encompass the vision-language domain \cite{kayser2021vil, li2018vqa, marasovic2020natural, park2018multimodal, wu2018faithful}. \citeauthor{kayser2022explaining} introduced the MIMIC-NLE dataset, derived from the MIMIC-CXR dataset \cite{johnson2019mimic}, to advance interpretability and accessibility in the context of chest X-ray analysis. This dataset is currently the only publicly available resource specifically designed for generating NLEs related to chest X-rays.

In their work, \citeauthor{kayser2022explaining} also introduced benchmark methods for generating explanations, such as DPT (DenseNet-121 \cite{huang2017densely} combined with GPT-2 \cite{radford2019language}) and RATCHET \cite{hou2021ratchet}. While GPT-2 has demonstrated effective performance in general domains \cite{kayser2021vil}, it shows limitations when applied to medical NLE generation due to its reliance on commonsense knowledge, which is often insufficient for specialized medical contexts. The DPT model, in particular, struggled with generating accurate explanations for chest X-ray images due to its dependency on non-specialized knowledge sources. The MIMIC-NLE dataset provides explanations for predicted pathologies, making it a comprehensive resource for evaluating the quality of NLEs in medical imaging. Our approach demonstrates a substantial improvement over previous methods, underscoring the value of integrating domain-specific knowledge through KG augmentation in the generation of NLEs. Also, \citeauthor{rio-torto2024parameterefficient} have investigated parameter-efficient training techniques for the NLE task.

\subsubsection{Vision-Language Models.}
The emergence of advanced Large Language Models (LLMs) like LLaMA \cite{touvron2023llama} and GPT-4 \cite{achiam2023gpt} has showcased significant improvements in text generation capabilities. Building on these developments, researchers have increasingly focused on extending these models to handle multimodal inputs, such as visual data. Despite these efforts, fully integrating visual and textual modalities remains a challenging endeavor, particularly in areas such as understanding spatial relationships, mathematical reasoning, and counting. \citeauthor{bordes2024introduction} categorize VLMs into four primary categories, contrastive training, masking, pre-trained backbones, and generative vision-language models.

VLMs in the pre-trained backbone category often utilize open-source LLMs, such as LLaMA/Viccuna, to learn mappings between a pre-trained image encoder and the LLM. This approach is computationally efficient, as it avoids training both text and image encoders from scratch, instead focusing on aligning the representations generated by these pre-trained models \cite{li2024llava, li2023blip, zhu2023minigpt}.

Our work falls within the category of VLMs based on pre-trained backbones. Specifically, we employ an LLaVA \cite{li2024llava} model which is based on pre-trained LLM alongside a pre-trained vision model, connected via a Projector that learns to map the visual information to the language model space. This approach leverages the strengths of pre-trained vision and language models.

Visual Instruction Tuning has been effectively employed in models like LLAVA \cite{li2024llava}, where a pre-trained LLM is trained on visual inputs provided by a vision model using carefully curated instruction datasets. This process typically involves mapping image features from pre-trained models like CLIP into the LLM's embedding space. While more sophisticated projection methods, such as lightweight transformers in CLIPCap \cite{mokady2021clipcap} and the Bert-based Q-Former in BLIP-2 \cite{li2023blip}, have been explored, the linear layer approach remains a popular choice due to its simplicity and efficiency.

\subsubsection{Knowledge Graph.}
A KG represents relationships between a set of entities, offering a structured approach to capturing and utilizing interconnected information. In addition to KG-TREAT \cite{liu2024kg} and MKG-FENN \cite{Wu_Sun_He_Chen_Luo_2024}, recent works such as ImgFact \cite{Liu_Zhang_Li_Wang_Li_Jiang_Jiang_Xiao_Chen_2024} and VCTP \cite{Chen_Zhou_Shen_Hong_Sun_Gutfreund_Gan_2024} have pushed the boundaries of KG integration. ImgFact introduces a large-scale multimodal KG with triplet fact grounding to improve visual reasoning tasks, while VCTP applies visual chain-of-thought prompting to facilitate knowledge-based visual reasoning. In the radiology domain, the RadGraph \cite{jain2021radgraph} method introduces a pioneering approach to constructing KGs from medical reports, providing a systematic means of extracting and representing clinically relevant information. RadGraph not only defines the methodology for creating these graphs but also supplies a publicly available dataset, which significantly aids research and development in this area.

Unlike these methods, our privacy-preserving KG-RAG framework focuses on medical NLE tasks. By leveraging clinically relevant relationships (e.g., ‘suggestive\_of’) and ensuring data privacy through de-identified triplets for NLE generation.

\subsubsection{Retrieval Augmented Generation.} RAG \cite{lewis2020retrieval} represents a significant advancement in enhancing language models beyond the capabilities achieved through traditional supervised fine-tuning. In RAG, external knowledge is dynamically retrieved and integrated into the language model, thereby augmenting its generative capabilities with new information that the model has not been explicitly trained on. This approach is particularly valuable when addressing dynamic or domain-specific knowledge, such as medical information, where up-to-date and specialized content is crucial.
However, as identified by \cite{zeng2024good}, the use of RAG in sensitive domains like healthcare raises substantial security concerns. Specifically, there is a risk that malicious actors could exploit RAG systems by prompting the model to retrieve and expose sensitive information from the retrieval datastore, potentially leading to privacy breaches. This vulnerability underscores the necessity for more secure implementations of RAG, particularly in the context of handling confidential medical data.

\section{Methodology}
\label{sec:guidelines}
In this paper, we present a novel approach for generating NLEs via KG-RAG for thoracic pathologies. Our methodology utilizes medical vision models in conjunction with large language models, resulting in three distinct frameworks: KG-LLaVA, Med-XPT, and Bio-LLaVA, as illustrated in Figure \ref{fig1}. Each framework is designed to leverage the strengths of KG-RAG in enhancing the accuracy and contextual relevance of NLEs in the medical domain.

\subsection{Pathology Classification}

For pathology classification, the input X-ray image \( X \in \mathbb{R}^{H \times W \times C} \) is processed by a visual encoder, such as MedCLIP or ViT-L/14, to generate visual feature embeddings:

\begin{equation}
Z_v = V(X),
\label{eq:image-embedding}
\end{equation}

\noindent where \( V(X) \) represents the visual encoder that maps \( X \) to a high-dimensional feature vector \( Z_v \in \mathbb{R}^{d_v} \).

The extracted visual features \( Z_v \) are then passed through a multi-layer perceptron (MLP) to classify pathologies:

\begin{equation}
\hat{y} = \text{MLP}(Z_v),
\end{equation}


\noindent where \( \hat{y} \in \mathbb{R}^p \) denotes the predicted probabilities for \( p \) pathology classes.

In addition to predicting the associated certainty levels along with the identified pathologies in X-ray images, we adopted a methodology based on the approach outlined by \cite{kayser2022explaining}. This method involves the prediction of 10 distinct pathologies, each categorized into three certainty levels—negative, uncertain, or positive—using the UMultiClass strategy introduced by \cite{irvin2019chexpert}.

The UMultiClass strategy applies a classification function \( f \) to map each pathology prediction \( \hat{y}_i \) to a certainty level \( c_i \in \{\text{negative, uncertain, positive}\} \) based on predefined thresholds:

\begin{equation}
c_i = f(\hat{y}_i) = 
\begin{cases}
\text{negative} & \text{if } \hat{y}_i < \theta_{\text{neg}}, \\
\text{uncertain} & \text{if } \theta_{\text{neg}} \leq \hat{y}_i < \theta_{\text{pos}}, \\
\text{positive} & \text{if } \hat{y}_i \geq \theta_{\text{pos}},
\end{cases}
\label{eq:certainty-classification}
\end{equation}

\noindent where \( \theta_{\text{neg}} \) and \( \theta_{\text{pos}} \) are the thresholds for classifying certainty levels.

We process the visual features \( Z_v \) extracted from the medical vision model through the MLP to enhance the model’s capability to interpret and classify the visual features of the X-ray images. This process improves the accuracy of both the pathology predictions \( \hat{y} \) and their corresponding certainty levels \( c_i \).

\subsection{Knowledge Graph Retrieval}

The Knowledge Graph (KG) is represented as a collection of triplets \((e_i, r_{ij}, e_j)\), where \( e_i \) and \( e_j \) are medical entities, and \( r_{ij} \) is the relationship between them extracted from clinical reports:

\begin{equation}
\text{KG} = \{(e_i, r_{ij}, e_j) \mid e_i, e_j \in \mathcal{E}, r_{ij} \in \mathcal{R}\}.
\label{eq:kg-definition}
\end{equation}

Each triplet is embedded into a high-dimensional feature space using the MedCLIP text encoder \( f_{KG} \):

\begin{equation}
Z_k = f_{KG}(e_i, r_{ij}, e_j),
\label{eq:kg-embedding}
\end{equation}

\noindent where \( Z_k \in \mathbb{R}^{d_k} \) represents the embedding of the triplet \( (e_i, r_{ij}, e_j) \).

To retrieve the most relevant triplets for a given query image \( X \), the image is first processed by the visual encoder to obtain the image embedding (refer to Equation~\ref{eq:image-embedding}).

The retrieval process is conducted by computing the cosine similarity between the image embedding \( Z_v \) and each triplet embedding \( Z_k \) in the datastore:

\begin{equation}
\text{sim}(Z_v, Z_k) = \frac{Z_v \cdot Z_k}{\|Z_v\| \|Z_k\|}.
\label{eq:cosine-similarity}
\end{equation}

The top-\(k\) most relevant triplets \( T_{\text{retrieved}} \) are then selected based on the highest similarity scores:

\begin{equation}
T_{\text{retrieved}} = \arg\max_{Z_k \in \text{KG}} \, \text{sim}(Z_v, Z_k).
\label{eq:top-k-retrieval}
\end{equation}

To address the privacy risks highlighted in Section 2, we propose a KG-based RAG approach as a secure alternative. Unlike traditional RAG systems, which may retrieve information traceable to specific patients, our approach employs a KG composed of general medical terms, entities, and relationships. This structured representation of knowledge graph abstracts away direct patient-specific details, significantly reducing the risk of inadvertently exposing sensitive information.

In our framework, we augment the model with knowledge instances retrieved from a constructed datastore. This method enhances the model's performance by providing relevant contextual knowledge while protecting sensitive medical data.
To enable effective knowledge retrieval, we constructed a datastore comprising KG triplets derived exclusively from the MIMIC-CXR training set \cite{johnson2019mimic} using the RadGraph model \cite{jain2021radgraph}. These triplets are aligned with the MIMIC-NLE \cite{kayser2022explaining} training set, ensuring no overlap with the dev and test sets, thereby preventing data leakage. We focus specifically on triplets with the ‘suggestive\_of’ relationship (e.g., “opacity suggestive\_of pneumonia”) as these triplets are more directly relevant to explaining the presence of pathologies. These triplet embeddings were generated using a medical CLIP model and are exclusively stored in the datastore, deliberately excluding any image features. This design facilitates a cross-modal retrieval process, where images can be used to query and retrieve relevant KG triplets.

KG-RAG retrieves contextually relevant knowledge by focusing on semantically similar information rather than exact matches. Even when minor deviations occur, the Vision-Language Model (VLM) leverages visual cues to maintain accurate pathology identification.

\section{Experiment}

\subsection{Dataset}

For our study, we utilized the MIMIC-NLE dataset \cite{kayser2022explaining}, which is derived from MIMIC-CXR chest X-ray dataset \cite{johnson2019mimic}. This is currently the only publically available dataset for chest X-ray NLEs. The MIMIC-NLE dataset includes diagnoses, evidence labels, and corresponding NLEs for those diagnoses. For a detailed description of the dataset, please refer to the comprehensive overview provided in \cite{kayser2022explaining}. The dataset consists of 38,003 NLEs and is divided into training, validation, and testing subsets, containing 37,016, 273, and 714 entries, respectively.

The visual instruction tuning approach is employed in both KG-LLaVA and Bio-LLaVA, where it enhances the model’s instruction-following abilities and generalization performance across various medical imaging tasks. However, instruction tuning was not applied to the Med-XPT framework, which focuses instead on leveraging its custom architecture for generating NLEs.

For a comprehensive description of the instruction-format dataset and the tailored prompts for pathology explanations, please refer to Appendix A in the supplementary material.

\subsection{Implementation Details}

\textbf{Projector.} The GPT-2 based model uses a transformer-based projector, while the LLaMA/Vicuna-based model uses an MLP-based projector. The projector aligns the image embeddings with the features required by the Language Model (LM). Specifically, the visual features \( Z_v \) are projected into the language model's embedding space using a trainable linear layer:

\begin{equation}
H_v = W_p \cdot Z_v + b_p,
\label{eq:projector}
\end{equation}

\noindent where \( W_p \in \mathbb{R}^{d_l \times d_v} \) is the weight matrix, \( b_p \in \mathbb{R}^{d_l} \) is the bias term, and \( H_v \in \mathbb{R}^{d_l} \) represents the projected visual features aligned with the language model's embedding space.

\textbf{Vision Models.} \label{formats}
In our proposed framework, we employ two distinct vision models tailored for different components of the system. First, we utilize the MedCLIP model \cite{wang2022medclip}, which serves dual purposes. It is integrated with the DPT \cite{kayser2022explaining} framework for NLE generation and is also instrumental in constructing the KGR datastore. MedCLIP's robust capabilities enable effective retrieval of relevant information based on image features, ensuring that the retrieved knowledge aligns with the visual content of medical images.

For the KGR process within our LLaVA-based framework, we again utilize MedCLIP to perform the retrieval itself based on the image features. However, for the extraction of visual features and their subsequent projection into the language embedding space, we employ the ViT-L/14 \cite{radford2021learning} CLIP model as the vision encoder. The ViT-L CLIP model extracts visual features from input images (refer to Equation~\ref{eq:image-embedding}).

These visual features are then projected into the language model's embedding space through a trainable projection matrix (refer to Equation~\ref{eq:projector}). This simple linear layer facilitates seamless integration with the language model. This dual-model approach allows us to leverage the specific strengths of both MedCLIP and ViT-L/14 within the framework, optimizing the generation of NLEs and ensuring accurate and contextually relevant retrieval of knowledge.

\textbf{Language Models.} 
For the decoding mechanism, we integrated language models such as GPT-2 \cite{radford2019language} and LLaMA/Viccuna \cite{touvron2023llama, peng2023instruction}, known for their effectiveness in natural language generation (NLG) tasks. The language model \( LM \) generates a probability distribution over target tokens \( Y = \{y_1, y_2, \ldots, y_n\} \) conditioned on the input sequence \( S \), which includes projected visual features, pathology labels, and retrieved knowledge triplets:

\begin{equation}
P(Y \mid S) = \prod_{i=1}^{n} P(y_i \mid S, y_1, y_2, \ldots, y_{i-1}).
\label{eq:conditional-probability}
\end{equation}

This structured prompt ensures that the generated NLEs are conditioned on relevant image features, pathology information, and retrieved knowledge. This process effectively combines visual data and textual information to produce accurate and contextually relevant NLEs.

\subsection{Training}
Our proposed frameworks, KG-LLaVA, Med-XPT, and Bio-LLaVA, each incorporate a KGR module to enhance their performance in generating NLEs for thoracic pathologies. KG-LLaVA builds upon the pre-trained LLaVA model, integrating the KGR module to leverage domain-specific knowledge derived from the input image. Med-XPT is trained from scratch, combining MedCLIP as the vision encoder with a transformer-based projector and GPT-2 as the language model, while incorporating the KG module to enrich the generation process. Bio-LLaVA adapts the LLaVA model by replacing the vision encoder with Bio-ViT-L and modifying the projection layer to accommodate the unique feature dimensions, also integrating the KG module to improve model performance. These frameworks were trained on the MIMIC-NLE dataset, allowing us to evaluate the effectiveness of the KGR module across different architectures.

\begin{table*}[t]

 \centering
 \scalebox{0.99}{
 \begin{tabular}{l|cclcc}
 \toprule
 Method&       AUC&B4&  MET.&R.L.&CIDEr\\
 \midrule
 RATCHET \cite{hou2021ratchet}&       66.4&4.7& 14.1
&22.2&37.9\\
 TieNet \cite{wang2018tienet}&        64.6&3.5& 12.4
&19.4&33.9\\
 DPT \cite{kayser2022explaining}&      62.5&2.4& 11.3
&15.4&17.4\\
 LoRA AE \cite{rio-torto2024parameterefficient} & 63.9& 4.0& \textbf{15.3}& 20.6&24.4\\
 Prompt AE \cite{rio-torto2024parameterefficient} & 61.3& 3.7& 14.4& 19.7&23.4\\
 Prefix AE \cite{rio-torto2024parameterefficient}& 65.2& 3.7& 14.7& 19.7&21.5\\
 LLaMA-Adapt AE \cite{rio-torto2024parameterefficient} & 63.9& 4.3& 14.6& 21.4&29.7\\
 + multi-modal \cite{rio-torto2024parameterefficient} & 64.9& 3.0& 14.1& 18.6&18.4\\
 + MSE loss \cite{rio-torto2024parameterefficient} & 62.3& 2.0& 10.0& 14.2&14.2\\
 \midrule
 KG-LLaVA&     \textbf{83.0}&\textbf{7.2}& 15.1&\textbf{25.0}&\textbf{62.2}\\
 \bottomrule
 \end{tabular}
 }
  \caption{Comparison of our KG-LLaVA with other baselines on the MIMIC-NLE test set, focusing on NLG metrics. The metrics include Area Under the Curve (AUC), BLEU-4 (B4), METEOR (MET.), ROUGE-L (R.L.), and CIDEr scores. 
  }
  \label{tab1}
 \end{table*}

\begin{table}[t]

 \centering
 \scalebox{0.99}{
 \begin{tabular}{c|cccc}
 \toprule
 Method&B4&  MET.&R.L.&CIDEr\\
 \midrule
 Bio-LLaVA& 5.7&  \underline{14.3}&\underline{23.0}& 46.7\\
 Med-XPT& \underline{7.0}& 11.0& 22.9& \textbf{62.7}\\
 KG-LLaVA& \textbf{7.2}& \textbf{15.1}&\textbf{25.0}& \underline{62.2}\\
 \bottomrule
 \end{tabular}
 }
  \caption{Performance comparison of our proposed frameworks—Bio-LLaVA, Med-XPT, and KG-LLaVA—on the MIMIC-NLE test set, focusing on NLG metrics. All frameworks incorporate KG-RAG module. Evaluation metrics include BLEU-4 (B4), METEOR (MET.), ROUGE-L (R.L.), and CIDEr, scores.
  }
  \label{tab2}
 \end{table}

Refer to Appendix B in the supplementary material for detailed implementation and training configurations of KG-LLaVA, Med-XPT, and Bio-LLaVA.

\subsection{Evaluation Metrics} In line with \cite{kayser2022explaining}, we evaluate NLEs only for correctly predicted labels. Prior research has indicated that standard automated natural language generation (NLG) metrics often fall short in assessing NLE quality due to the variability in expressing similar meanings using different syntactic structures and semantic interpretations \cite{kayser2021vil}. For our evaluation, we report the widely used NLG metrics: BLEU, ROUGE, CIDEr, and METEOR \cite{kayser2021vil}.

Among these, CIDEr is particularly relevant as it measures the n-gram overlap with ground-truth explanations, placing emphasis on contextual relevance. By retrieving pathology-specific triplets through KG-RAG, our approach increases the likelihood of incorporating relevant medical terms into the generated NLEs. This contributes to the higher CIDEr scores observed in our results, highlighting the effectiveness of KG-RAG in enhancing the contextual accuracy of explanations.

\section{Results and Discussions}

\subsection{Comparison with Other Methods} In this study, we evaluated the performance of our proposed KG-LLaVA framework against several well-established methods, including RATCHET \cite{hou2021ratchet}, TieNet \cite{wang2018tienet}, and DPT \cite{kayser2022explaining}, using the MIMIC-NLE \cite{kayser2022explaining} dataset. The results, as summarized in Table \ref{tab1}, clearly demonstrate that KG-LLaVA outperforms the previous methods across a range of evaluation metrics.

KG-LLaVA achieves an AUC of 83.0, which is significantly higher than the AUC scores reported for RATCHET (66.4), TieNet (64.6), and DPT (62.5). This substantial improvement underscores the effectiveness of our approach in accurately classifying and generating relevant explanations for thoracic pathologies. Additionally, KG-LLaVA excels in key NLG metrics, including BLEU-4 (7.2), ROUGE-L (25.0) and CIDEr (62.2), highlighting its ability to generate high-quality, contextually accurate explanations.

Notably, while KG-LLaVA slightly outperforms RATCHET \cite{hou2021ratchet} in METEOR (15.1 vs. 14.1), the overall superior performance of KG-LLaVA across the other metrics underscores the strength of incorporating KG-RAG module into a vision-language framework. These results suggest that KG-LLaVA has the potential to set a new benchmark for generating NLEs in medical imaging tasks, particularly for diagnosing thoracic pathologies.

While \cite{rio-torto2024parameterefficient} methods focused on optimizing model parameters, our approach leveraged the KG-RAG module and effective use of LoRA for fine-tuning, achieving superior performance without compromising the model's complexity or parameter efficiency. This makes KG-LLaVA not only the best-performing model but also a robust and scalable solution for medical NLE generation.

\begin{table}[t]

\centering
\resizebox{0.95\columnwidth}{!}{
\begin{tabular}{c|ccccc} 
\toprule
Methods&  RAG&B4& METEOR &R-L&CIDEr \\ 
\midrule
Med-XPT&  -&2.0& 7.8& 12.8& 17.4\\ 
KG-LLaVA&  -&7.0& 15.0& 24.4& 60.1\\ 
\midrule
Med-XPT& NLE& 6.7& 13.5& 22.2& 59.3\\ 
KG-LLaVA& NLE& 6.8& 15.0& 24.6& 58.8\\ 
\midrule
Med-XPT& KG& 7.0& 11.0& 22.9& \textbf{62.7}\\
KG-LLaVA& KG& \textbf{7.2}& \textbf{15.1}& \textbf{25.0}& 62.2\\
\bottomrule
\end{tabular}
}
\caption{Comparative analysis of the performance of Med-XPT, and KG-LLaVA across different RAG methods and without any RAG. The table includes results for NLG metrics such as BLEU-4 (B4), METEOR, ROUGE-L (R-L), and CIDEr. The "-" row shows results without any RAG integration, the "NLE" row represents results with natural language explanation-based RAG, and the "KG" row reflects the performance when the knowledge graph retrieval module is used.}
\label{tab3}
\end{table}

\subsection{Comparison of different LLMs} We further assessed the performance of our three proposed frameworks—Bio-LLaVA, Med-XPT, and KG-LLaVA all of which incorporate the KG-RAG module. The results, detailed in Table \ref{tab2}, provide insights into the comparative strengths of each framework in generating NLEs for thoracic pathologies.

KG-LLaVA demonstrates the highest overall performance, leading in BLEU-4 (7.2), METEOR (15.1), and ROUGE-L (25.0). These results reflect its superior ability to generate accurate and contextually rich explanations. Med-XPT, on the other hand, performs exceptionally well in CIDEr (62.7), indicating its effectiveness in capturing the diversity and richness of language necessary for high-quality NLEs. Bio-LLaVA, while slightly behind in some metrics, still shows strong performance in METEOR (14.3) and ROUGE-L (23.0).
These findings underscore the flexibility and efficacy of the KG-RAG module across different architectures. The variability in performance across the frameworks suggests that the choice of architecture can be optimized based on specific aspects of the NLE task, such as accuracy, linguistic richness, or diversity in the generated explanations.

\subsection{Impact of Different RAG Methods} We conducted a detailed comparison of the two frameworks—Med-XPT and KG-LLaVA—across various configurations: without any Retrieval Augmented Generation (RAG), with standard NLE, and with our proposed KG Retrieval module. The results, as shown in Table \ref{tab3}, illustrate the impact of different RAG methods on the performance of these frameworks in generating accurate and contextually rich NLEs for thoracic pathologies.

In the NLE configuration, where standard NLEs are generated without KG enhancement, both Med-XPT and KG-LLaVA exhibit strong performance, with KG-LLaVA slightly leading in most metrics. This indicates that while both frameworks leverage their respective architectures effectively, the pre-training knowledge embedded in KG-LLaVA likely contributes to its superior performance.

Additional discussion is provided in Appendix C of the supplementary material.

\subsection{Role of Instructional Prompts}
An ablation study on the role of instructional prompts is available in Appendix D.

\begin{figure}[t]
\centering
\includegraphics[width=0.99\columnwidth]{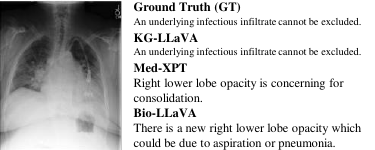} 
\caption{Comparison of NLEs generated by different models—KG-LLaVA, Med-XPT, and Bio-LLaVA—against the ground truth (GT) for a specific thoracic pathology case. The image depicts a chest X-ray used as input, with the corresponding NLEs. KG-LLaVA accurately matches the GT by identifying the underlying abnormalities, while Bio-LLaVA and Med-XPT offer alternative interpretations, reflecting the models' varying strengths and limitations in clinical reasoning.}
\label{fig2}
\end{figure} 

\subsection{Qualitative Results}
The qualitative analysis of the generated NLEs from our proposed frameworks—KG-LLaVA, Bio-LLaVA, and Med-XPT—highlights distinct differences in their alignment with the ground truth (GT) as shown in Figure \ref{fig2}. KG-LLaVA accurately replicates the GT by identifying the underlying infectious infiltrate, showcasing its strong alignment with expert annotations. In contrast, Bio-LLaVA introduces an alternative diagnosis, suggesting a new right lower lobe opacity possibly due to aspiration or pneumonia, which, while clinically plausible, diverges from the GT. Med-XPT incorrectly focuses on a right lower lobe opacity concerning consolidation, indicating challenges in precise localization and consistency. These findings underscore KG-LLaVA's effectiveness in generating accurate NLEs, while also illustrating the flexibility and limitations of Bio-LLaVA and Med-XPT in clinical interpretation.

Overall, these results highlight the significant impact of the KG-RAG module on improving model performance across different architectures. KG-LLaVA consistently shows strong results across all configurations, underscoring its potential as a leading framework for generating precise and contextually relevant NLEs in the medical imaging domain. The findings also suggest that the choice of RAG method plays a crucial role in determining the quality of NLEs, with KG-RAG offering the most substantial benefits including data security.

\subsection{Analysis of Failure Case}
A detailed discussion of failure cases is provided in Appendix E.
\subsection{Limitation}
Our framework requires a KG model like RadGraph for constructing triplets. For new modalities, practitioners may need to adapt or train alternative graph-processing models. While our design is intended for research purposes, further clinical validation is necessary before real-world deployment. 

\section{Conclusion}
In this paper, we introduced a novel approach for generating NLEs for thoracic pathologies by integrating the KG-RAG module into vision-language models. Our KG-RAG effectively enhances the accuracy and contextual relevance of NLEs by incorporating domain-specific knowledge. Evaluated across three distinct frameworks—KG-LLaVA, Med-XPT, and Bio-LLaVA—our method consistently outperformed established models like RATCHET, TieNet, and DPT on the MIMIC-NLE dataset, highlighting the robustness and versatility of the KG-RAG approach.

Moreover, the inclusion of the KG-RAG module addresses critical privacy concerns by abstracting patient-specific details, thereby safeguarding data security and preventing data leakage. These findings underscore the critical role of integrating domain-specific knowledge in advancing vision-language models for medical imaging while ensuring the security and privacy of sensitive medical data. This approach sets a new benchmark for AI-driven diagnostics, paving the way for more transparent, accurate, and trustworthy healthcare systems.

\section*{Acknowledgments}
This work was supported in part by the Institute of Information and Communications Technology Planning and Evaluation (IITP) Grant funded by the Korea Government (MSIT) under Grant 2022-0-00078 (Explainable Logical Reasoning for Medical Knowledge Generation), Grant IITP-2024-RS-2023-00258649, Grant RS-2022-00155911, Grant 2021-0-02068 and by the National Research Foundation of Korea (NRF) Grant funded by the Korea Government (MSIT) under Grant RS-2024-00334321.

\bigskip

\bibliography{aaai25}

\bigskip
\clearpage

\twocolumn[{
    \begin{center}
        {\LARGE \textbf{Supplementary Material}}\\[1em]
        {\LARGE \textbf{LLaVA Needs More Knowledge: Retrieval Augmented Natural Language Generation with Knowledge Graph for Explaining Thoracic Pathologies}}
        \vspace{1em}
    \end{center}
}]

\appendix

\section{Dataset and Instruction Prompts}
For our study, we constructed an instruction-format dataset based on the LLaVA \cite{li2024llava} framework, with specific modifications to the prompts. These modifications were designed to tailor the questions to elicit detailed explanations from the model regarding the reasoning behind the occurrence of specific pathologies observed in the images. The following questions were incorporated into our prompts to guide the model in generating explanations:
\textit{Which signs show that the patient has {pathologies}?,
Explain why these {pathologies} are present in the image?,
What evidence in the image indicates {pathologies}?,
How can you tell that the patient has {pathologies} from the image?,
What features suggest the presence of {pathologies} in this image?.}
These tailored prompts were crucial in enhancing the model's ability to provide reasoning for the presence of specific pathologies based on the visual evidence present in the images.

\section{Implementation Details}
\subsection{Training}
The KGR process is powered by the MedCLIP model, which projects the input image into a shared latent space. This projected representation is then used to retrieve corresponding triplets from the KG using the FAISS library \cite{johnson2019billion}, which employs a k-nearest neighbors algorithm to identify the most relevant triplets. These retrieved triplets serve as supplementary information, enriching the input to the language model and improving the model's ability to generate accurate and contextually relevant explanations.

Our training methodology across these frameworks—KG-LLaVA, Med-XPT, and Bio-LLaVA—underscores the flexibility and effectiveness of the KGR module, which consistently improves model performance by leveraging structured, domain-specific knowledge in both vision-language tasks and NLE generation.

\subsection{Training Setup}
We trained the LLaVA \cite{li2024llava} model with the integrated KGR module on the MIMIC-NLE dataset. The training process involved fine-tuning the model to maximize the likelihood of generating high-quality NLEs using the additional knowledge provided by the KG.

The training was conducted with LoRA \cite{hu2021lora} (Low-Rank Adaptation) for training efficiency. We used the pre-trained LLaVA \cite{li2024llava} model as the baseline, with a learning rate of 2e-4, and employed the cosine learning rate scheduler. The training was performed over 5 epochs with a batch size of 8 per device and a gradient accumulation of 2 steps. The vision encoder used was CLIP ViT-L \cite{radford2021learning}, and the multimodal projector was an MLP with two layers using the GELU activation function.

We also employed techniques such as gradient checkpointing and mixed precision to optimize memory usage and training speed. The maximum sequence length for the model was set to 2048 tokens to accommodate the large inputs from both text and image modalities. The model was trained on the MIMIC-NLE dataset, and the training data was preprocessed lazily to streamline the process.

In addition to our KG-LLaVA framework, we also applied our KGR module to Med-XPT architecture, which utilizes MedCLIP as the vision encoder, a transformer-based projector, and GPT-2 as the language model. We followed a similar structured training approach. We again used the FAISS \cite{johnson2019billion} library for retrieval based on image features. The training script included several key parameters: the visual encoder (medclip-vit) was used to retrieve captions stored in a JSON file, with the retrieval results incorporated into the input prompts. The attention mechanism within the model was tuned using a cross-attention size of 7, and we enabled the training of the decoder along with the attention mechanism. The training process spanned 15 epochs with a learning rate of 1e-4 and a batch size of 8, utilizing gradient accumulation with one step. The retrieved captions and templates were dynamically integrated into the training pipeline, which facilitated the model's ability to generate accurate NLEs based on the retrieved knowledge.

Lastly, for the Bio-LLaVA framework, which incorporates the Bio-ViT-L model as the vision encoder, we customized the projection layer to handle the unique feature dimensions of Bio-ViT-L. The training of Bio-LLaVA followed the same structured methodology, emphasizing the integration of KG triplets to enhance model performance.

\section{Impact of Different RAG Methods}
Despite the significantly smaller size and less extensive training of the GPT-2 language model compared to LLaVA, Med-XPT—which utilizes GPT-2—achieves competitive performance, particularly when augmented with the KG and NLE-based RAG methods. This suggests that integrating domain-specific knowledge can effectively mitigate the limitations of smaller models, enabling them to generate high-quality explanations.

Notably, when comparing Med-XPT to the DPT \cite{kayser2022explaining} framework, which also uses GPT-2 but without the benefit of these advanced RAG methods, we observe a substantial improvement in performance. DPT’s lower scores across the board highlight the critical role that the KG and NLE-based RAG play in enhancing the model's explanatory capabilities. This further emphasizes that even smaller models, when equipped with the right augmentation techniques, can perform on par with larger, more complex models like LLaVA.

The most significant improvements are observed in the KG configuration, where the KGR module is employed. KG-LLaVA leads in BLEU-4, ROUGE-L, and METEOR metrics, while Med-XPT excels in CIDEr scores. This demonstrates the effectiveness of the KG-RAG module in enhancing the richness and contextual relevance of the generated explanations, particularly in KG-LLaVA.

In addition to the strong performance metrics, it's important to highlight that KG-LLaVA addresses critical privacy concerns by abstracting patient-specific details through a KG-based datastore. While KG-LLaVA may have slightly lower scores on some metrics compared to other models, its ability to safeguard data security and prevent data leakage makes it a valuable solution in medical AI applications, offering a robust balance between performance and privacy preservation.

\section{Ablation Study on Instruction Prompts}

To evaluate the role of instructional prompts in guiding the quality of Natural Language Explanations (NLEs), we conducted an ablation study on 20 randomly selected test instances from the dataset. Our study analyzed the impact of different prompt configurations on BLEU-4, ROUGE-L, and CIDEr scores, which measure precision, recall, and contextual relevance, respectively.

We considered three configurations:

\begin{itemize}
    \item \textbf{Original Prompt (Full Set of Questions)}: The complete set of instructional questions used in the baseline configuration.
    \item \textbf{Additional Broad Question}: Adding the general question, “What patterns in the image are associated with specific pathologies?”
    \item \textbf{No Instructional Questions}: All instructional prompts were removed.
\end{itemize}

The results of the ablation study are summarized in Table~\ref{tab:instructional_ablation}.

\begin{table}[t]
\centering

\caption{Ablation Study Results for Instructional Prompts}
\label{tab:instructional_ablation}
\begin{tabular}{c|ccc}
\toprule
\textbf{Prompt Configuration}       & \textbf{B-4} & \textbf{R-L} & \textbf{CIDEr} \\
\midrule
\textbf{Original Prompt} (Full Set) & 7.56          & 25.33         & 82.59        \\
\textbf{Additional Broad Question}  & 5.24          & 24.62         & 63.24       \\
\textbf{No Instructional Questions} & 6.70          & 25.52         & 70.77       \\
\bottomrule
\end{tabular}
\end{table}

\subsection*{Key Observations}

\begin{itemize}
    \item \textbf{Original Prompt (Full Set of Questions)}:  
    The baseline configuration achieved the highest scores across all metrics (BLEU-4: 7.56, ROUGE-L: 25.33, CIDEr: 82.59). This indicates that a comprehensive set of instructional questions helps produce structured, coherent, and clinically relevant explanations.

    \item \textbf{Additional Broad Question (“What patterns in the image are associated with specific pathologies?”)}:  
    Adding a broader question reduced BLEU-4 (5.24) and CIDEr (63.24), while ROUGE-L (24.62) remained close to the baseline. This suggests that overly general questions introduce ambiguity, leading to less precise and less targeted explanations.

    \item \textbf{No Instructional Questions}:  
    Removing all instructional questions significantly reduced performance, with BLEU-4 at 6.70 and CIDEr at 70.77, though ROUGE-L (25.52) stayed near the baseline. This demonstrates that instructional prompts are crucial for guiding the model towards specific, relevant, and structured outputs.
\end{itemize}

The ablation study highlights the importance of instructional prompts in enhancing NLE quality. Well-designed prompts focused on specific diagnostic features yield the best performance. Omitting prompts or introducing overly broad questions reduces explanation quality across precision, recall, and contextual relevance. These findings reinforce the role of instructional prompts as a critical component in our framework for generating accurate and coherent explanations.

\section{Failure Case Analysis}
\label{sec:failure_cases}

In this section, we present five failure cases where the KG-LLaVA model demonstrated the lowest performance in terms of CIDEr and ROUGE-L scores. Each case highlights specific challenges faced by the model, such as ambiguity in retrieved triplets, misinterpretation of image features, and difficulty in localizing abnormalities accurately. These insights provide a deeper understanding of the model's limitations and suggest pathways for improvement.

\begin{table*}[htbp]
\centering
\caption{Case 1: Misinterpretation of Opacity Location}
\label{tab:misinterpretation_opacity}
\renewcommand{\arraystretch}{1.5} 
\begin{tabular}{p{3cm} | p{13cm}}
\toprule

\textbf{Input Image} & \includegraphics[width=0.2\textwidth]{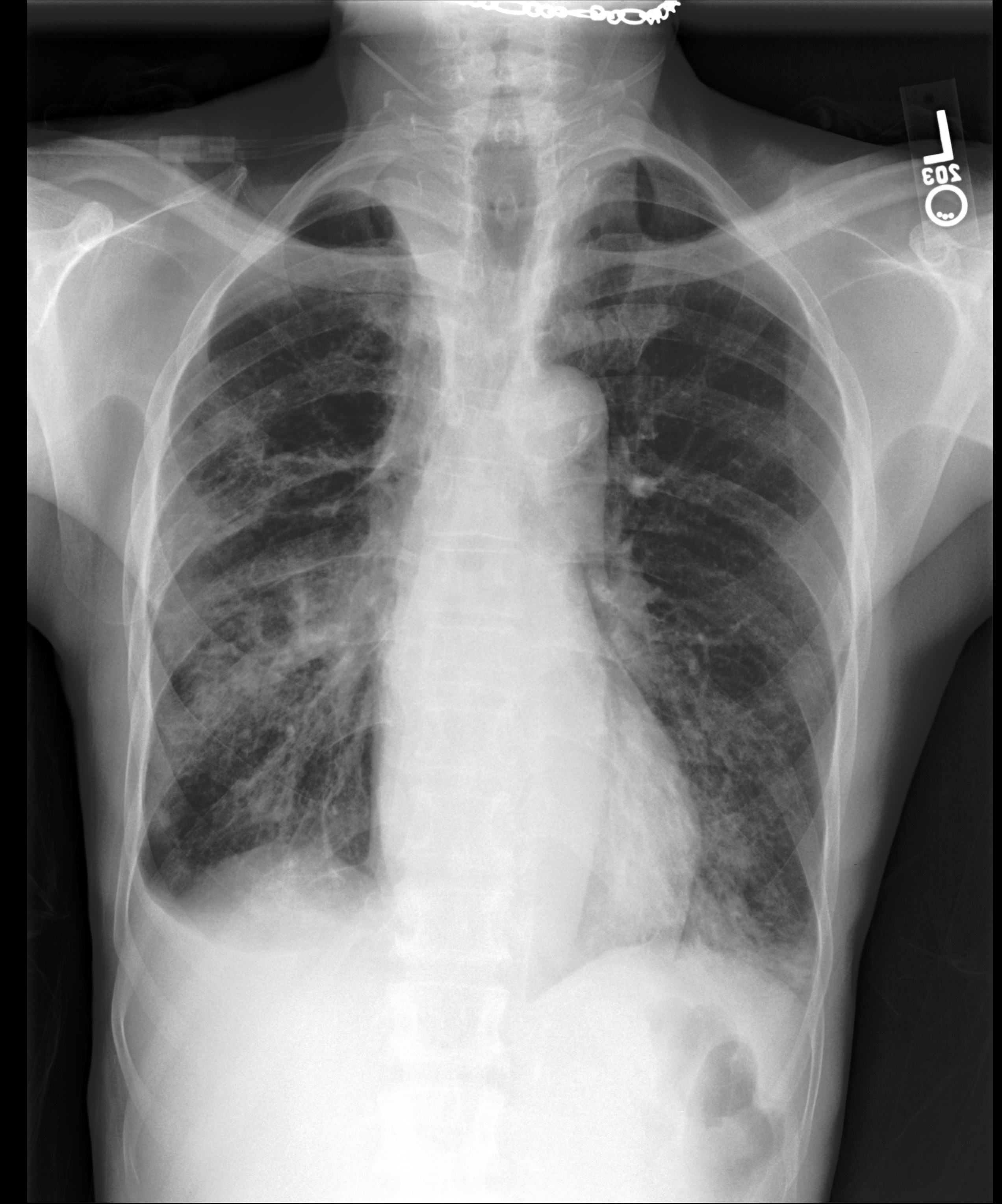} \\ \midrule

\textbf{Triplets} & \textit{opacities suggestive\_of effusions; opacities suggestive\_of effusions; opacities suggestive\_of effusions; opacities suggestive\_of effusions; opacities suggestive\_of effusions} \\

\textbf{Question} & \textit{Which signs show that the patient has positive Lung Lesion, positive Lung Opacity, positive Pleural Effusion, positive Pneumonia?} \\

\textbf{Prediction} & \textit{There is a new opacity in the right lower lobe with a small amount of pleural fluid tracking along the lateral margin of the lung.} \\

\textbf{Ground Truth (GT)} & \textit{New bibasilar patchy airspace opacities, concerning for aspiration pneumonia.} \\

\bottomrule
\end{tabular}
\end{table*}

\begin{table*}[htbp]
\centering
\caption{Case 2: Ambiguity in Diagnosing Edema and Effusion}
\label{tab:ambiguity_edema_effusion}
\renewcommand{\arraystretch}{1.5} 
\begin{tabular}{p{3cm} | p{13cm}}
\toprule

\textbf{Input Image} & \includegraphics[width=0.2\textwidth]{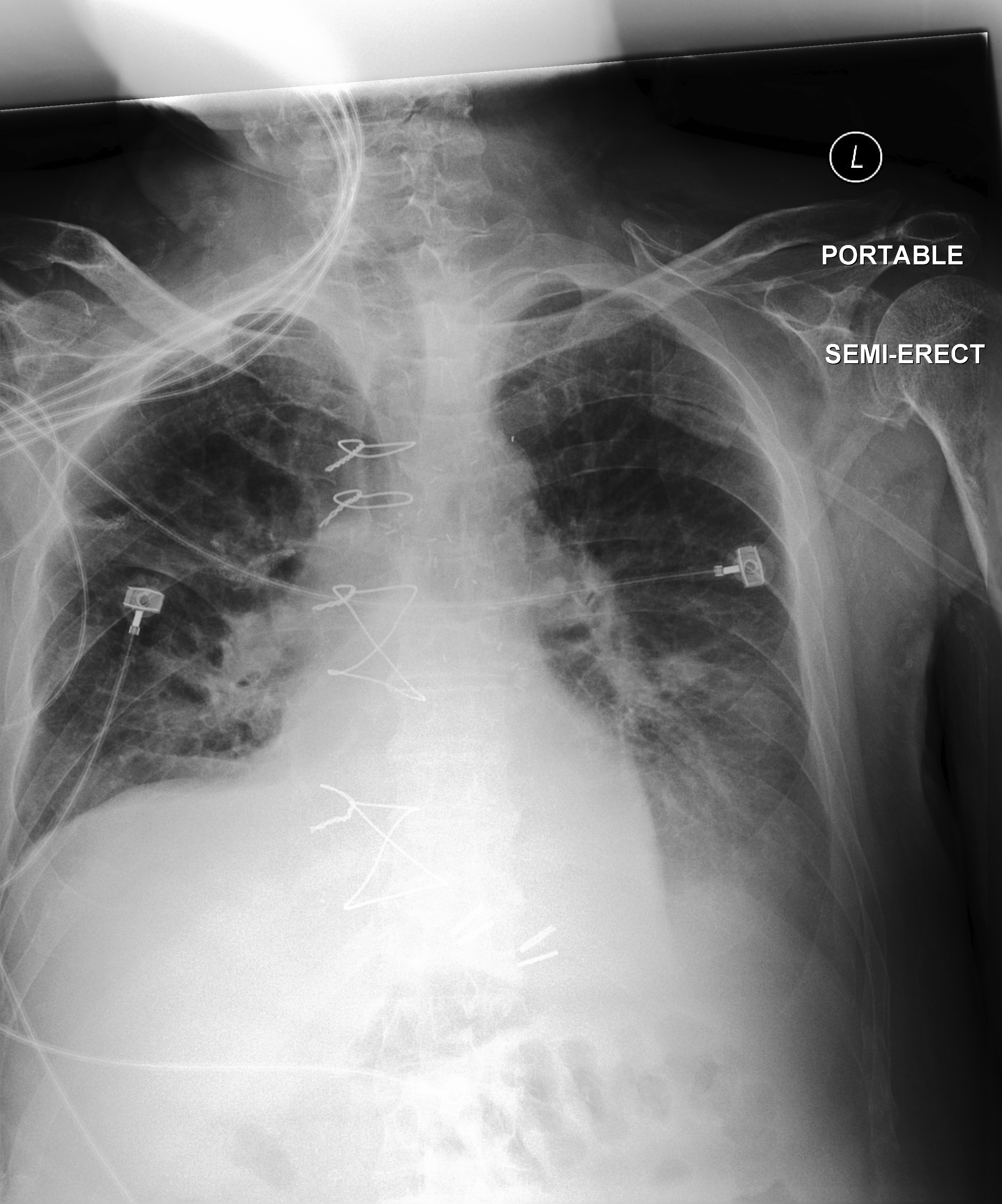} \\ \midrule

\textbf{Triplets} & \textit{radiopacity suggestive\_of edema; opacities suggestive\_of effusions; opacities suggestive\_of effusions; opacities suggestive\_of effusions; opacities suggestive\_of effusions} \\

\textbf{Question} & \textit{Explain why these positive Atelectasis, uncertain Edema, positive Lung Opacity, positive Pleural Effusion are present in the image.} \\

\textbf{Prediction} & \textit{Bibasilar opacities are likely atelectasis.} \\

\textbf{Ground Truth (GT)} & \textit{A hazy opacity over the left lung base suggests a layering pleural effusion.} \\

\bottomrule
\end{tabular}
\end{table*}

\begin{table*}[htbp]
\centering
\caption{Case 3: Inconsistent Interpretation of Pleural Effusion}
\label{tab:inconsistent_pleural_effusion}
\renewcommand{\arraystretch}{1.5} 
\begin{tabular}{p{3cm} | p{13cm}}
\toprule

\textbf{Input Image} & \includegraphics[width=0.2\textwidth]{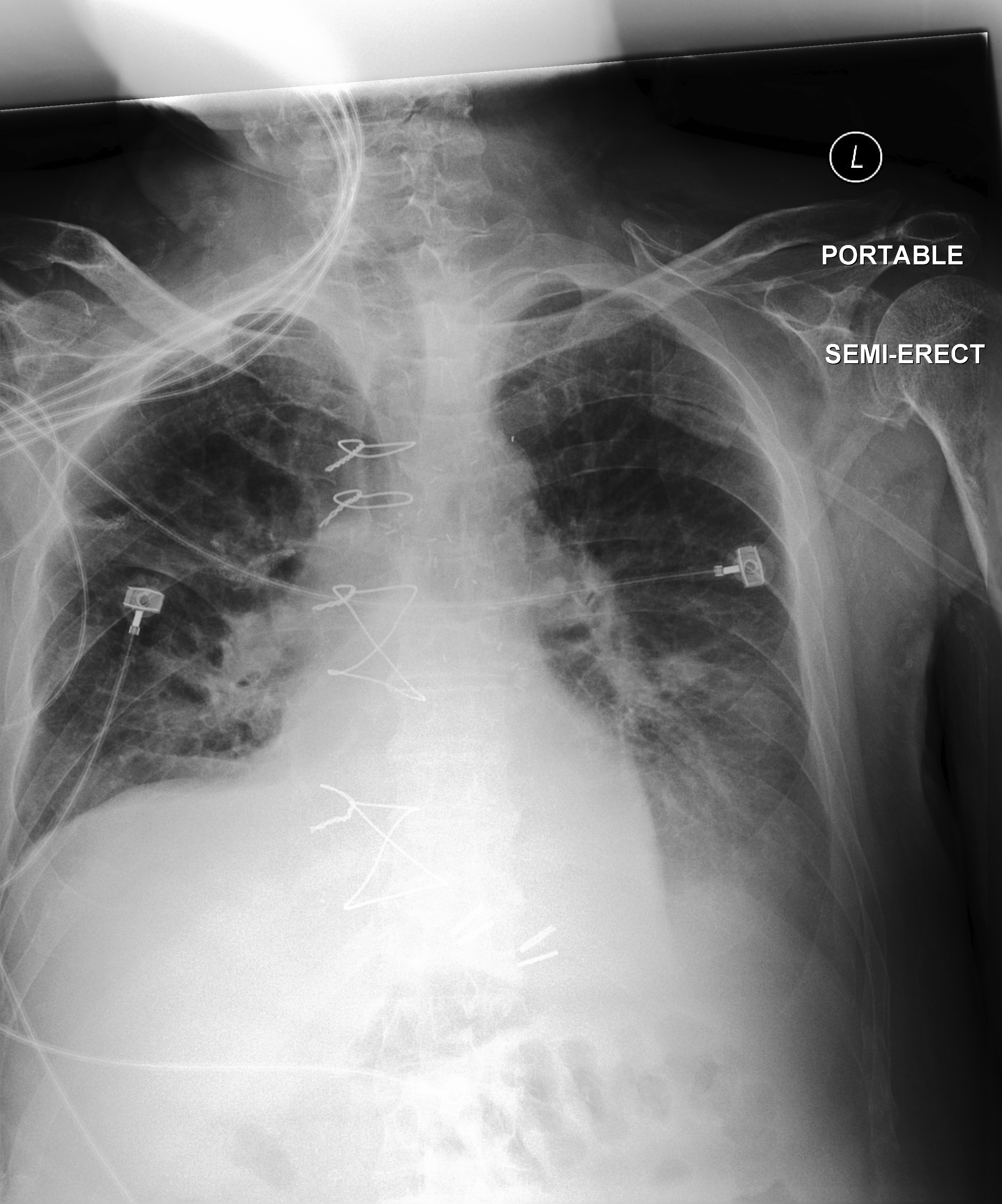} \\ \midrule

\textbf{Triplets} & \textit{radiopacity suggestive\_of edema; opacities suggestive\_of effusions; opacities suggestive\_of effusions; opacities suggestive\_of effusions; opacities suggestive\_of effusions} \\

\textbf{Question} & \textit{Explain why these positive Atelectasis, uncertain Edema, positive Lung Opacity, uncertain Pleural Effusion are present in the image.} \\

\textbf{Prediction} & \textit{Bibasilar opacities, likely atelectasis.} \\

\textbf{Ground Truth (GT)} & \textit{A small area of hazy opacity at the right costophrenic angle may represent a small layering pleural effusion.} \\

\bottomrule
\end{tabular}
\end{table*}

\begin{table*}[htbp]
\centering
\caption{Case 4: Limited Context for Multifocal Pneumonia}
\label{tab:limited_multifocal_pneumonia}
\renewcommand{\arraystretch}{1.5} 
\begin{tabular}{p{3cm} | p{13cm}}
\toprule

\textbf{Input Image} & \includegraphics[width=0.2\textwidth]{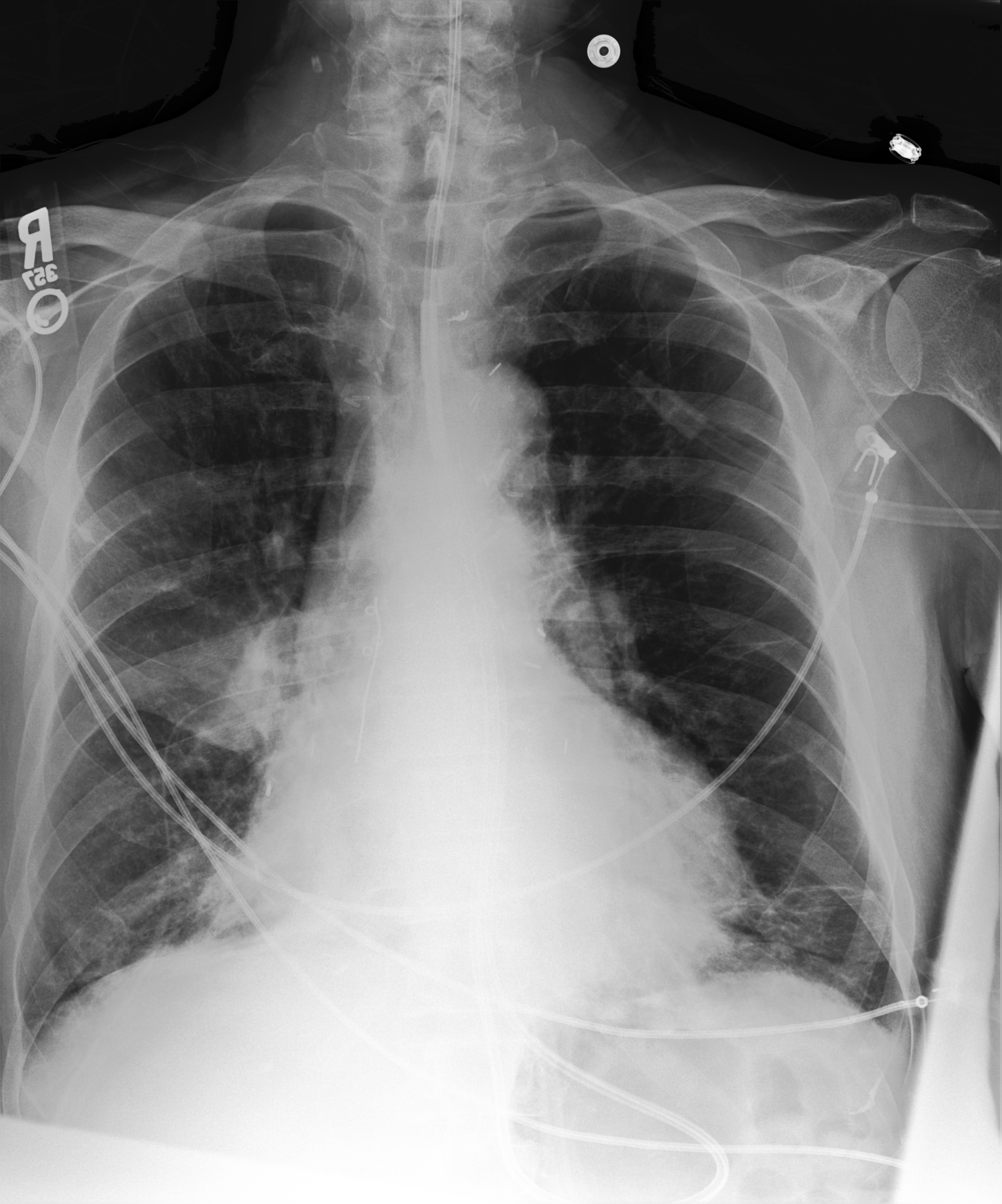} \\ \midrule

\textbf{Triplets} & \textit{line suggestive\_of hydropneumothorax; Tubes suggestive\_of drain; opacification suggestive\_of consolidation; opacification suggestive\_of consolidation; opacification suggestive\_of consolidation} \\

\textbf{Question} & \textit{What evidence in the image indicates positive Lung Opacity, positive Pleural Effusion, positive Pneumonia?} \\

\textbf{Prediction} & \textit{There is a new opacity in the right lower lobe which is concerning for pneumonia.} \\

\textbf{Ground Truth (GT)} & \textit{Multiple opacifications likely represent multifocal pneumonia, possibly due to aspiration.} \\

\bottomrule
\end{tabular}
\end{table*}

\begin{table*}[htbp]
\centering
\caption{Case 5: Confusion Between Atelectasis and Effusion}
\label{tab:confusion_atelectasis_effusion}
\renewcommand{\arraystretch}{1.5} 
\begin{tabular}{p{3cm} | p{13cm}}
\toprule

\textbf{Input Image} & \includegraphics[width=0.2\textwidth]{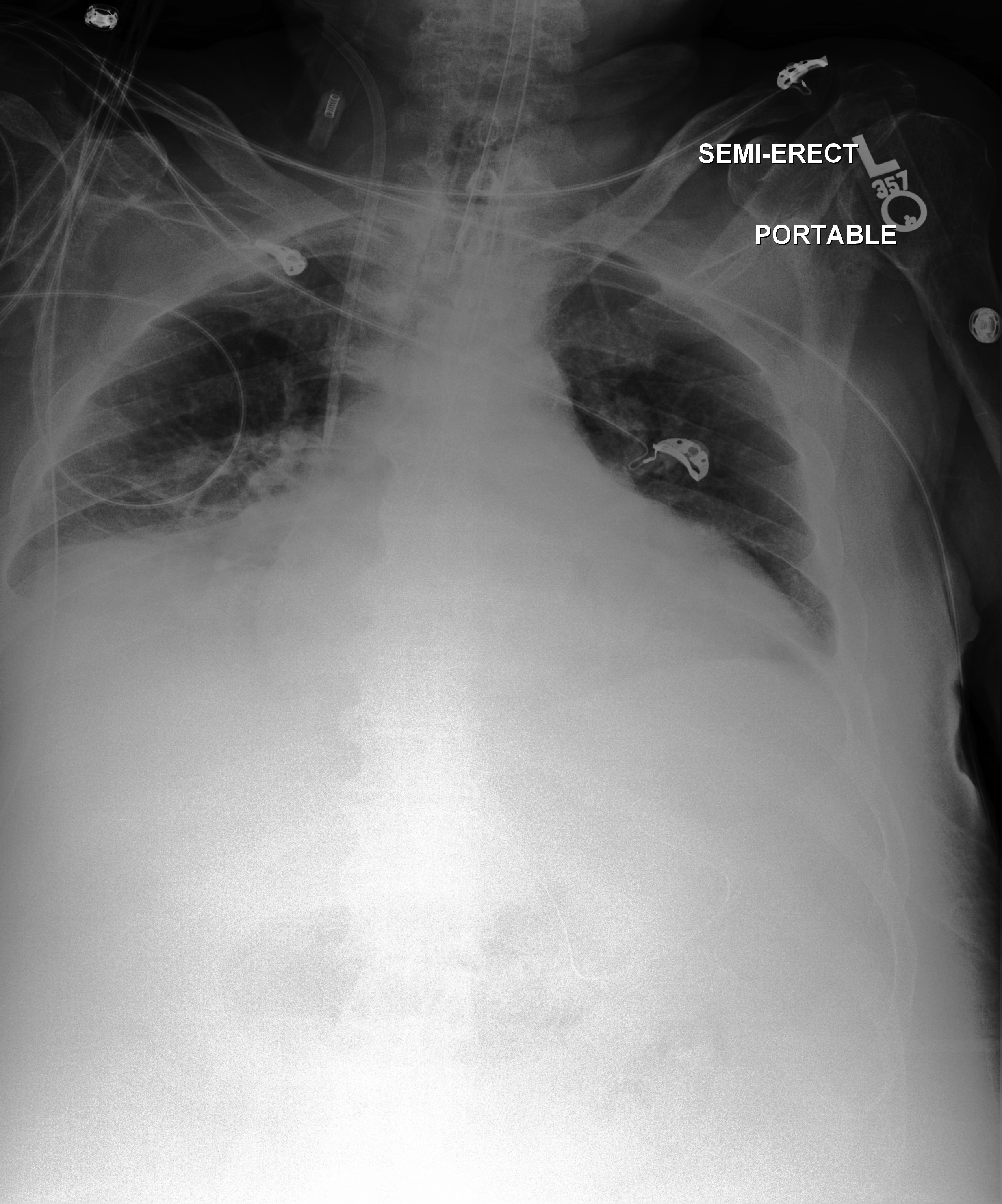} \\ \midrule

\textbf{Triplets} & \textit{atelectasis suggestive\_of scarring; atelectasis suggestive\_of scarring; Tubes suggestive\_of drain; opacification suggestive\_of collapse; opacification suggestive\_of collapse} \\

\textbf{Question} & \textit{What features suggest the presence of positive Atelectasis, positive Lung Opacity, positive Pleural Effusion in this image?} \\

\textbf{Prediction} & \textit{There is a left retrocardiac opacity and likely bilateral pleural effusions.} \\

\textbf{Ground Truth (GT)} & \textit{Substantial bibasilar opacification can be explained by atelectasis.} \\

\bottomrule
\end{tabular}
\end{table*}

Case 1 presented in Table \ref{tab:misinterpretation_opacity}, the model focused on a \textit{right lower lobe opacity}, whereas the ground truth describes \textit{bibasilar opacities}. The repeated and generic triplets (\textit{“opacities suggestive of effusions”}) provided insufficient contextual specificity. The model’s failure to capture the bilateral nature of the opacities led to an incorrect interpretation of the location, reducing explanation accuracy. This highlights the need for more precise and diverse triplets to improve spatial reasoning.

Case 2 presented in Table \ref{tab:ambiguity_edema_effusion} shows that the model defaulted to \textit{atelectasis} as an explanation, likely influenced by the ambiguity in the retrieved triplets, which mention both \textit{“radiopacity suggestive of edema”} and \textit{“opacities suggestive of effusions”}. The overlap between findings related to atelectasis and pleural effusion caused the model to miss the correct interpretation. More discriminative triplets distinguishing between these pathologies could mitigate this issue.

Case 3 presented in Table \ref{tab:inconsistent_pleural_effusion} shows that the model leaned towards \textit{atelectasis} rather than \textit{pleural effusion}. The lack of specificity in the retrieved triplets led to an incorrect explanation. This case illustrates the model's difficulty in handling subtle differences between overlapping radiological findings, emphasizing the need for triplets that capture nuanced distinctions.

Case 4 presented in Table \ref{tab:limited_multifocal_pneumonia} shows that the model's prediction missed the \textit{multifocal} nature of the pneumonia. The retrieved triplets emphasized \textit{“consolidation”} but lacked context about the distribution pattern. This case highlights the need for triplets that describe multifocal or distributed findings to guide more accurate explanations.


Case 5 presented in Table \ref{tab:confusion_atelectasis_effusion} shows that the model incorrectly identified \textit{bilateral pleural effusions} instead of \textit{atelectasis}. The triplets retrieved (\textit{“opacification suggestive of collapse”}) did not sufficiently differentiate between collapse and effusion. Enhancing the triplet selection process to provide clearer distinctions between these findings could improve the model's accuracy.

\subsection{Summary of Insights}

The failure cases reveal several key challenges:
\begin{itemize}
    \item \textbf{Ambiguity in Triplets}: Repeated or generic triplets introduce uncertainty, causing the model to default to common diagnoses.
    \item \textbf{Localization Errors}: The model struggles to correctly identify the location of abnormalities when multiple findings are present.
    \item \textbf{Lack of Context for Complex Diagnoses}: Triplets do not always capture multifocal or distributed patterns, limiting the model’s ability to handle complex cases.
\end{itemize}

Future improvements can focus on refining the knowledge graph with more specific and context-aware triplets, enhancing spatial reasoning capabilities, and incorporating multi-hop reasoning to improve diagnostic accuracy.

\section{Effect of the Number of Retrieved Knowledge Graph Triplets (K) on NLE}

\begin{table}[t]
\caption{Effect of the Number of Retrieved Knowledge Graph Triplets (K) on NLE Performance. The evaluation metrics include BLEU-4 (B4), METEOR, ROUGE-L (R-L), and CIDEr.}
\label{tab4}
\centering
\scalebox{0.99}{
\begin{tabular}{c|cccc}
\toprule
\textbf{K} & \textbf{B4} & \textbf{METEOR} & \textbf{R-L} & \textbf{CIDEr} \\
\midrule
1 & 7.3 & 15.4 & 25.3 & 60.6 \\
3 & 7.1 & 15.3 & 25.0 & 58.2 \\
5 & 7.2 & 15.3 & 24.9 & 58.6 \\
7 & 7.2 & 15.1 & 25.0 & 62.2 \\
\bottomrule
\end{tabular}
}
\end{table}

We conducted an ablation study to investigate the impact of varying the number of retrieved Knowledge Graph (KG) triplets \(K\) on the quality of the generated Natural Language Explanations (NLEs). Table~\ref{tab4} presents the evaluation results across BLEU-4 (B4), METEOR, ROUGE-L (R-L), and CIDEr scores.

When retrieving a single triplet (\(K=1\)), the model achieves strong baseline performance (BLEU-4: 7.3, METEOR: 15.4, ROUGE-L: 25.3, CIDEr: 60.6). This indicates that even a minimal amount of retrieved knowledge can provide sufficient context for generating coherent and relevant explanations.

As \(K\) increases to 3 and 5, the performance remains consistent, with minor fluctuations in BLEU-4 (7.1 to 7.2) and METEOR (15.3). The ROUGE-L scores exhibit a slight decline (25.0 to 24.9), suggesting that while additional triplets enrich the context, they do not significantly improve recall-based measures. The CIDEr scores for \(K=3\) (58.2) and \(K=5\) (58.6) reflect a minor drop compared to \(K=1\), indicating that excessive information can introduce noise, potentially reducing the precision of the explanations.

When \(K\) is increased to 7, the CIDEr score peaks at 62.2, demonstrating that a higher number of retrieved triplets can enhance contextual relevance and diversity, particularly for complex cases. This improvement in CIDEr suggests that retrieving more knowledge can increase the likelihood of including relevant terms and concepts in the generated explanations. However, the BLEU-4 (7.2) and METEOR (15.1) scores remain stable, indicating that beyond a certain threshold, additional triplets contribute more to contextual richness than to syntactic precision.

The results in Table~\ref{tab4} highlight a trade-off between the quantity of retrieved knowledge and the quality of NLEs. While retrieving more triplets (\(K=7\)) can improve contextual relevance, as evidenced by the higher CIDEr score, the benefits plateau for precision and recall-based metrics like BLEU-4 and ROUGE-L. Therefore, selecting an optimal \(K\) depends on balancing contextual richness with the risk of introducing noise. For our KG-LLaVA model, \(K=7\) provides the best overall balance, enhancing the model’s ability to generate informative and contextually relevant explanations.

\begin{table}[t]
\caption{Performance Comparison of Uni-modal and Cross-modal Retrieval on NLE Performance - This table compares the performance of the KG-LLaVA model when using Uni-modal versus Cross-modal retrieval methods for generating Natural Language Explanations (NLEs).}\label{tab5}
\centering
\scalebox{0.99}

\begin{tabular}{c|cccc}  
\toprule
K & B4& METEOR & R-L& CIDEr  \\
 \midrule
 Uni-modal& 5.8& 14.4& 23.2&49.9\\
 Cross-modal& 7.2& 15.1& 25.0&62.2\\
\bottomrule
\end{tabular}

\end{table}

\section{Comparison of Uni-modal and Cross-modal Retrieval on NLE Performance}

The Cross-modal retrieval method demonstrates a clear advantage over the Uni-modal approach across all evaluation metrics as shown in Table \ref{tab5}, with a particularly notable improvement in CIDEr, where the score increases from 49.9 to 62.2. This substantial gain highlights the effectiveness of integrating both visual and textual modalities in the retrieval process, allowing the model to generate more contextually relevant and accurate Natural Language Explanations (NLEs). The Cross-modal approach enhances the model's ability to interpret complex medical images, leading to higher-quality explanations that are more aligned with clinical expectations.

In contrast, the Uni-modal approach, which retrieves triplets by matching the input image with similar images in the datastore, shows comparatively lower performance. This image-to-image retrieval method, while effective, does not fully leverage the multimodal capabilities that Cross-modal retrieval offers. Moreover, storing images in the datastore raises significant privacy concerns, as it involves retaining patient-specific visual data, which could potentially be traced back to individual patients.

Our method, by abstracting patient-specific details and operating directly within the latent space for image-to-text retrieval, significantly mitigates these privacy risks. This approach not only enhances performance but also aligns with stringent data privacy requirements, making it particularly suitable for clinical applications where data security is paramount. The ability to achieve superior results without compromising on privacy underscores the versatility and practicality of our KG-RAG framework, reinforcing its potential for broader adoption in healthcare settings. This experiment clearly validates the privacy-preserving design of our method, which is a critical contribution outlined in our work.
\end{document}